\pdfoutput=1

\documentclass[11pt]{article}

\usepackage[]{ACL2023}

\usepackage{times}
\usepackage{latexsym}
\usepackage{graphicx}
\usepackage{subfigure}
\usepackage{CJKutf8}
\usepackage{makecell}
\usepackage[T1]{fontenc}

\usepackage[utf8]{inputenc}

\usepackage{microtype}

\usepackage{inconsolata}
\usepackage{booktabs}

\title{A New Dataset and Empirical Study for Sentence Simplification in Chinese}

\renewcommand\footnotemark{}
\author{\textbf{Shiping Yang\textsuperscript{*\dag}\thanks{\textsuperscript{*} Equal contribution}
\thanks{\textsuperscript{\dag} Work done during internship}, Renliang Sun\textsuperscript{*}, Xiaojun Wan} \\
         Wangxuan Institute of Computer Technology, Peking University\\
         Center for Data Science, Peking University\\
         The MOE Key Laboratory of Computational Linguistics, Peking University\\
  \texttt{yangshipingnlp@gmail.com} \\ \texttt{sunrenliang@stu.pku.edu.cn} \\
  \texttt{wanxiaojun@pku.edu.cn} \\
}

\begin{document}
\maketitle
\begin{abstract}
Sentence Simplification is a valuable technique that can benefit language learners and children a lot. However, current research focuses more on English sentence simplification. 
The development of Chinese sentence simplification is relatively slow due to the lack of data. 
To alleviate this limitation, this paper introduces CSS, a new dataset for assessing sentence simplification in Chinese. We collect manual simplifications from human annotators and perform data analysis to show the difference between English and Chinese sentence simplifications.
Furthermore, we test several unsupervised and zero/few-shot learning methods on CSS and analyze the automatic evaluation and human evaluation results. 
In the end, we explore whether Large Language Models can serve as high-quality Chinese sentence simplification systems by evaluating them on CSS.
\end{abstract}

\section{Introduction}
Sentence Simplification (SS) is the task of modifying a sentence to make it easier to understand and improve its accessibility for a wider audience, while retaining most of its original meaning \cite{alva2020data}. 
Automatic SS systems provide reading assistance to children \cite{de2010text,kajiwara2013selecting}, non-native readers \cite{paetzold2016unsupervised}, and people with reading disabilities \cite{carroll1998practical,rello2013dyswebxia,evans2014evaluation}.

Currently, there are multiple datasets for English sentence simplification to choose from, such as WikiSmall \cite{zhu2010monolingual}, WikiLarge \cite{wikilarge} and Newsela \cite{newsela}. However, few simplification-specific datasets are available for other languages.
Research on other popular languages like Spanish \cite{vstajner2015automatic,saggion2015making}, French \cite{gala2020alector}, Russian \cite{RuEval,RussianWithLearnerdata} and Italian \cite{brunato2015design,tonelli2016simpitiki} has gained momentum in recent years. Unfortunately, research on Chinese sentence simplification is still scarce: to the best of our knowledge, there is currently no publicly available simplification corpora for training, even lacking a dataset to evaluate the ability of simplification models. These limitations greatly prevent the development of Chinese sentence simplification.

Creating a parallel dataset for Chinese sentence simplification poses considerable challenges.
Most datasets for SS use the automatic sentence alignment method to construct \cite{wikilarge,RussianWithLearnerdata}, which relies on the existence of large-scale simplification corpus like Simple English Wikipedia\footnote{\url{https://dumps.wikimedia.org/simplewiki}}.
However, there are no suitable data sources in Chinese.
Another possible option is to translate the English dataset to Chinese through neural machine translation \cite{RuEval,li2022c3kg}. Nevertheless, Chinese is different from English in both grammatical structure and language habits, which leads to a significant difference in text simplification between these two languages.
Simplifying sentences with the help of human experts is also an option, like it was done in Newsela or ALECTOR\footnote{ALECTOR is a SS dataset for evaluation in French.}\cite{gala2020alector}, but this is expensive and slow.
Due to the above reasons, we decided to manually construct a dataset only for evaluation, achieving a trade-off between cost and feasibility. 

In this study, we annotate and propose CSS (\textbf{C}hinese \textbf{S}entence \textbf{S}implification dataset), a new Chinese dataset for the evaluation of SS models. We then apply several unsupervised SS methods and zero/few-shot learning methods on the dataset and analyze the advantages and disadvantages of the methods for Chinese sentence simplification. Furthermore, we study the behavior of popular metrics for English SS when using them to evaluate simplifications produced by Chinese SS systems. 


We are committed to promoting research on Chinese sentence simplification. In general, our contributions can be summarized as follows:

\begin{itemize}
\item We create a high-quality dataset named CSS for the evaluation of Chinese SS models. We will publicly release CSS at \url{https://github.com/maybenotime/CSS}, it will be the first open-source simplification-specific dataset in Chinese.
\item We conduct data analysis to compare the characteristics of CSS with English datasets, pointing out the difference between Chinese and English sentence simplification tasks.

\item We report the performance of several unsupervised methods and zero-shot/few-shot learning methods on our dataset, which could serve as the baselines for future studies.
\end{itemize} 

\section{Related Work}

\subsection{Sentence Simplification}   

Sentence simplification research has achieved promising progress in recent years. EditNTS \cite{dong2019editnts} simplifies a sentence with iterative explicit editing. ACCESS \cite{ACCESS} performs controllable simplification by conditioning specific control tokens. \citet{tst} proposed a simple and efficient simplification system based on sequence Tagging. However, these methods all relied on supervised parallel training corpora.

To overcome the scarcity of parallel SS corpus in low-resource languages, recent research has proposed many unsupervised methods to train simplification models without a labeled simplification corpus \cite{kajiwara2018text,surya2019unsupervised,katsuta2019,aprosio,kumariterative}. MUSS \cite{muss} obtains strong performance in French and Spanish, even outperforming the supervised state of the art. \citet{NMT} further improved the performance by building pseudo-SS corpora with an unsupervised approach. Finally, the experiments of multi-task learning \cite{dmitrieva2021multi} and cross-lingual learning \cite{mallinson2020zero} in sentence simplification shows the possibility of performing zero- and few-shot simplification without any parallel data, driving us to explore the Chinese SS task in the zero- and few-shot setting.

\subsection{Simplification Datasets in Multiple Languages}
There exist a lot of supervised training corpora \cite{newsela,wikilarge} and high-quality test datasets \cite{xu2016sari,hsplit,asset} for English SS. 
However, automatic SS systems in other popular languages also have extensive demand and application values. Researchers attempted to explore simplification in other languages \cite{aluisio2008towards,saggion2015making,kajiwara2018text}, but are limited by the lack of parallel corpora.

Recently, some works have focused on building SS datasets in other low-resource languages \cite{brunato2015design,battisti2020corpus,RuEval,RussianWithLearnerdata} to facilitate the development of multilingual SS techniques, such as ALECTOR \cite{gala2020alector}, Simpitiki \cite{tonelli2016simpitiki}, and Spanish part of Newsela \cite{newsela}. However, to our best knowledge, there is no work attempting to build a Chinese SS dataset, which hinders the development of Chinese SS systems.

\section{CSS}

In this section, we describe detailed information about our CSS dataset. 
Specifically, we first give the annotation process of the CSS dataset in Section~\ref{Data Collection}. 
Then, we show statistical information about our CSS dataset in Section~\ref{operation statistics}. 
In Sections~\ref{text features} and \ref{human rating}, we do automatic and manual data analysis on CSS. And an additional dataset for few-shot setting is described in Section ~\ref{additional}.

\subsection{Data Collection and Annotation} \label{Data Collection}
To obtain the raw texts for CSS, we randomly sampled original sentences from the PFR Chinese corpus.
Then, we asked the annotators who have passed the qualification test to simplify those sentences.
Except for manual simplifications, annotators were also asked to give the rewriting transformations they performed on the original sentences.
We introduce the PFR corpus and Preprocessing details in Appendix \ref{datasource}.



\paragraph{Operations Defined}
According to the previous studies on human simplification \cite{petersen2007text,feng2008text}, we define 4 simplification operations that can be performed in CSS: (1) lexical simplification (replacing complex words with synonyms or explaining idioms with a short sentence). (2) sentence splitting. (3) compression (deleting unimportant information). (4) sentence paraphrasing (word reordering or syntactic transformations).

\paragraph{Worker Requirements}

The requirements for workers are as follows: (1) native speakers of Chinese; (2) had education experience in university with at least a bachelor's degree; (3) passed the corresponding Qualification Test designed for our task (more details below). 
These requirements were designed to ensure the workers have a proficient level of Chinese, and are capable of performing the simplification task.

\paragraph{Qualification Test}
We designed a Qualification Test (QT) to measure the worker's simplification ability. 
The content of the test consisted of simplification operation recognition, specific sentence simplification, and free simplification. 
Before the QT, we showed them detailed explanations of the sentence simplification task and examples of multiple simplification operations we defined. 
After the annotators took the QT, all submissions were manually checked to filter out workers who could not perform the task correctly. 
We had 10 candidates take the QT, out of which 5 passed the test and entered the following annotation stage.

\paragraph{Annotation Round}
Workers who passed the QT would have access to this round. 
In addition to the simplification of each sentence, workers were also asked to annotate the simplification operations they performed on sentences and submit confidence scores (similar to ASSET \cite{asset}) on their simplifications using a five-point Likert scale (1:No Simplification Implemented, 2:Very Low, 5:Very High). 
We finally collected two simplified counterparts for each original sentence from different workers to fit the scenario with multiple transformations. Thus, our dataset is suitable for automatic evaluation metrics that require multiple references, such as BLEU \cite{bleu} and SARI \cite{xu2016sari}.

\begin{table*}[htbp]
\centering
\renewcommand{\arraystretch}{1.2}{
\begin{tabular}{ll}
\hline
Original &  \begin{CJK*}{UTF8}{gbsn}\footnotesize{\makecell[l]{五台县位于山西省忻州市，因境内有五台山而得名，而五台山位居全国四大佛教名山之首。\\Wutai County, located in Xinzhou City, Shanxi Province, is named after the Wutai Mountains,\\ which ranks first among the four most famous Buddhist mountains in China.}}\end{CJK*}\\ \hline
Reference & \begin{CJK*}{UTF8}{gbsn}\footnotesize{\makecell[l]{五台县在山西省忻州市，因为它里面有五台山从而得到这个名字。\\五台山在全国四大佛教名山中排名第一。\\Wutai County is in Xinzhou City, Shanxi Province, and got this name because of the Wutai Mountain.\\ Wutai Mountain ranks first among the four most famous Buddhist mountains in China.}}\end{CJK*} \\ \hline
Operations &  Lexical simplification; Sentence splitting; Sentence paraphrasing\\ \hline
\end{tabular}}
\caption{Simplification example with corresponding translations and operation tags in CSS.}
\label{table1}
\end{table*}

\paragraph{Simplification Guide}
Before the QT and the Annotation Round, workers were asked to read the task guide about how to do sentence simplification. 
We provided examples and definitions of lexical simplification, sentence splitting, compression, and sentence paraphrasing. 
We also included an example where all transformations were performed. 
To stimulate their creativity and motivation, we informed workers that they can simplify original sentences with which type of simplification operations at their discretion.
Additionally, we added bonuses to encourage workers to perform multiple transformations in a sentence.

\paragraph{Quality Control}
We added some fake examples to ensure the quality of the dataset. 
The fake examples were assigned to every worker in each annotation round.
We checked whether the workers gave reasonable simplifications by comparing their simplification results with the golden reference.
Besides, we manually checked the instance (an original-simplified sentence pair) with a confidence score of 1 and removed the original sentences that have no need to simplify from the dataset.

Table \ref{table1} presents an example of simplifications in CSS, together with corresponding translation and operation tags. Please refer to Appendix \ref{more_examples} for more examples.

\begin{table}[htbp]
\centering
\begin{tabular}{lccc}
\hline
 & CSS & ASSET & HSplit \\ 
\hline
Ori. Sentences & 383 & 359 & 359 \\
Num. of Ref. & 2 & 10 & 4 \\
Multi Operations & $\surd$ & $\surd$ & $\times$ \\
Operation Tag & $\surd$ & $\times$ & $\times$ \\ 
Tokens per Ref. & 47.29 & 19.04 & 25.49 \\ 
\hline
\end{tabular}
\caption{Basic statistics of CSS compared with ASSET and HSplit. From here on, we only report the statistics of the test set of ASSET for a fair comparison.}
\label{table2}
\end{table}

\begin{table*}[htbp]
\begin{tabular}{l|llll}
\hline
Operation & Lexical simplification & Sentence splitting & Compression & 
Sentence paraphrasing \\
\hline
Percentage(\%) & \multicolumn{1}{c}{91\%} & \multicolumn{1}{c}{20\%} & \multicolumn{1}{c}{45\%} & \multicolumn{1}{c}{60\%} \\
\hline
\end{tabular}
\caption{the percentage of each simplification operation applied to the original sentences, which is calculated by the number of the original sentences that have a certain operation occurring in references divided by the number of the original sentences in CSS. To some extent, this value can indicate which simplification operation is applicable to most sentences.}
\label{table3}
\end{table*}

\subsection{Statistics} \label{operation statistics}
CSS consists of 766 human simplifications associated with the 383 original sentences from the PFR corpus (two simplifications per original sentence). 
Table \ref{table2} presents basic statistics of CSS. 
We also show the same statistics of two mainstream English SS datasets for comparison.

Compared with previous English SS datasets, CSS offers fewer references for each original sentence but with rich rewriting transformations and additional information. 
While HSplit \cite{hsplit} contains simplifications produced mostly by sentence splitting, simplifications in CSS are more similar to ASSET's which involve multiple types of rewriting transformations.
CSS also provides additional simplification operation tags to show which types of rewriting transformations have been performed on this sentence, different from other datasets. 
Operation tags can provide help in the evaluation of controlled SS systems.


Table \ref{table3} shows the percentage of each simplification operation applied to the original sentences. 
It can be seen that most of the sentences in Chinese can be simplified by lexical simplification, and few annotators tended to simplify a sentence by sentence splitting. 
Compression and sentence paraphrasing are also common ways for Chinese SS.



\subsection{Dataset Analysis} \label{text features}
We further study the simplifications collected for CSS through a series of surface and syntax-based features. 

\begin{itemize}
\item[$\bullet$] \textbf{Number of simplification operations:}
The number of simplification operations on the simplification instance. 

\item[$\bullet$] \textbf{Number of sentence splits:}
The number of sentences in the simplification minus the number of sentences in the original sentence. 

\item[$\bullet$] \textbf{Compression level:}
The number of characters in the simplification divided by the number of characters in the original sentence.

\item[$\bullet$] \textbf{Replace-only Levenshtein distance:}
We report the Replace-only Levenshtein distance as described in ASSET, which is computed as character-level Levenshtein distance \cite{levenshtein1966binary} only with replace operations divided by the length of the shortest string. Therefore, this feature serves as a proxy for lexical simplification.

\item[$\bullet$] \textbf{Proportion of words deleted, added and reordered:} Number of words deleted/reordered\footnote{A reordered word is a word that is contained in the original sentence and simplification but not in the longest common subsequence.} from the original sentence divided by the number of words in the original sentence; and the number of words that were added to the original sentence divided by the number of words in the simplification.

\item[$\bullet$] \textbf{Word deletion only:}
A boolean feature shows whether the simplification is obtained only by deleting words from the original sentence. This feature captures extractive compression.

\item[$\bullet$] \textbf{Lexical complexity score ratio:}
Word ranks (in a frequency table) have been shown to be the best indicator of word complexity \cite{paetzold2016semeval}. We compute the log of word ranks as log-ranks, and then obtain the lexical complexity score\footnote{There is a difference in the computing way of lexical complexity score between the source code of \textit{tseval} and description in ASSET\cite{asset}, we following the version of the original paper.} by computing the mean squared log-ranks of words in a sentence (without stopwords). We use the Chinese common words frequency table, released by BLCU Corpus Center\footnote{\url{http://bcc.blcu.edu.cn/}}. The ratio is then the lexical complexity score on the simplification divided by that of the original sentence.

\item[$\bullet$] \textbf{Dependency tree depth ratio:}
We compute the ratio of the depth of the dependency parse tree of the simplification relative to that of the original sentence. When a simplification contains more than one sentence, we use the maximum depth of all dependency trees as the depth of the simplification. This feature is a good indicator to show the simplification in sentence structure.

\end{itemize}

\begin{table}[]
\centering
\begin{tabular}{lcc}
\hline
 & \multicolumn{1}{l}{CSS} & \multicolumn{1}{l}{ASSET} \\ \hline
Sentence Splitting & 11.8\% & 20.2\% \\
Compression(<75\%) & 9.1\% & 31.2\% \\
Word Reordering & 17.6\% & 28.3\% \\
Word Deletion Only & 5.6\% & 4.5\% \\
\hline
\end{tabular}
\caption{The percentage of sentences that: have at least one sentence split, have a compression level of 75\% or lower, have at least one reordered word, and operate word deletion only.}
\label{table4}
\end{table}

\begin{figure*}[t]
	\centering
	\begin{minipage}{0.24\linewidth}
		\centering
		\includegraphics[width=1.1\linewidth]{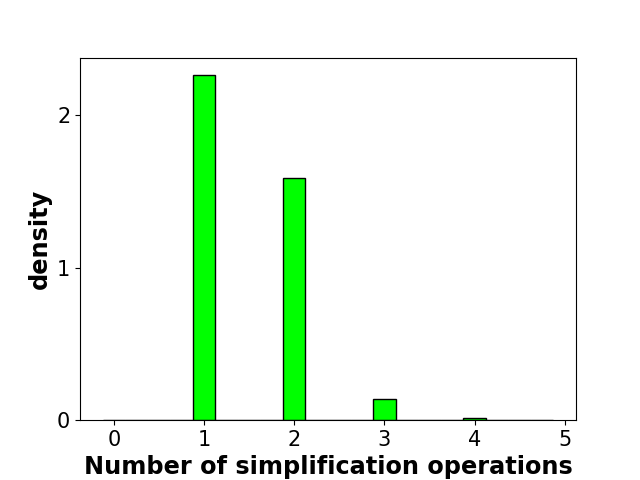}
	\end{minipage}
	\begin{minipage}{0.24\linewidth}
		\centering
		\includegraphics[width=1.1\linewidth]{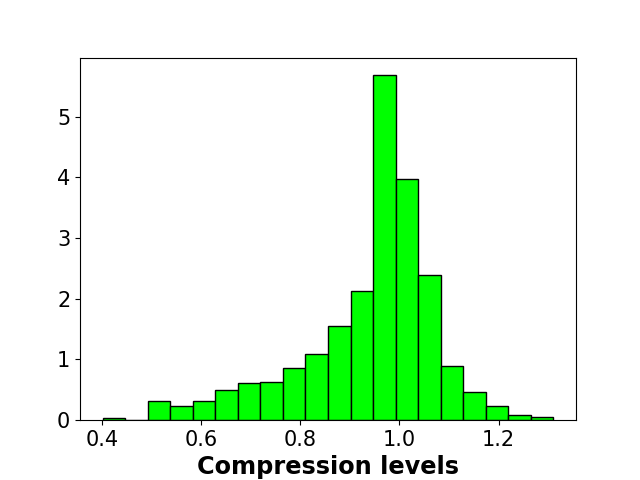}
	\end{minipage}
        \begin{minipage}{0.24\linewidth}
            \centering
            \includegraphics[width=1.1\linewidth]{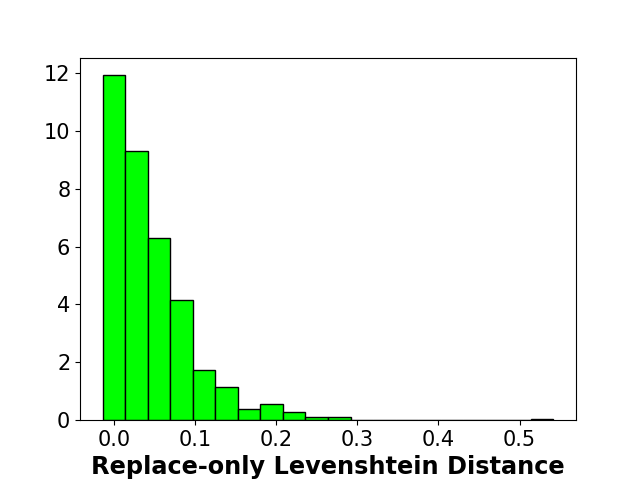}
        \end{minipage}
        \begin{minipage}{0.24\linewidth}
            \centering
            \includegraphics[width=1.1\linewidth]{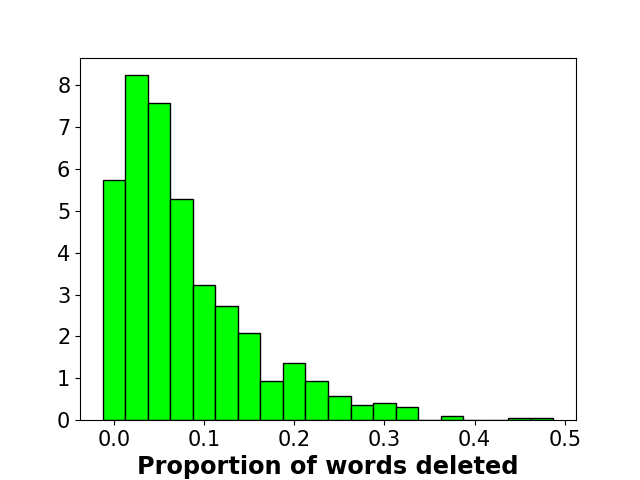}
        \end{minipage}
        \qquad
        \begin{minipage}{0.24\linewidth}
		\centering
		\includegraphics[width=1.1\linewidth]{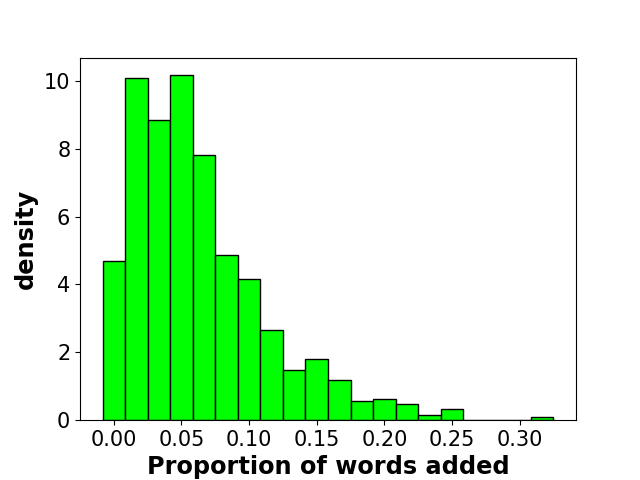}
	\end{minipage}
	\begin{minipage}{0.24\linewidth}
		\centering
		\includegraphics[width=1.1\linewidth]{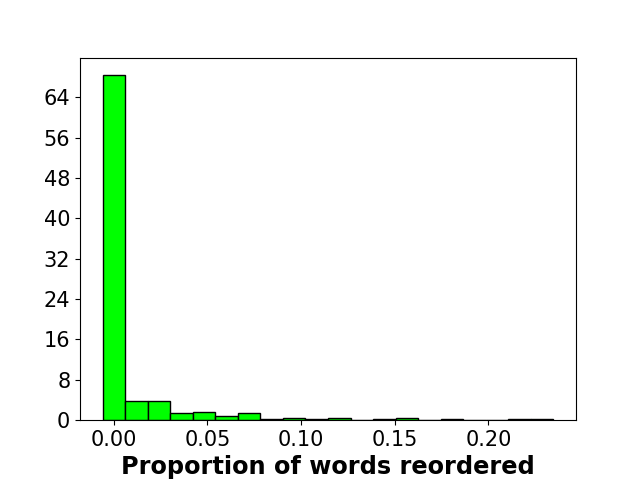}
	\end{minipage}
        \begin{minipage}{0.24\linewidth}
            \centering
            \includegraphics[width=1.1\linewidth]{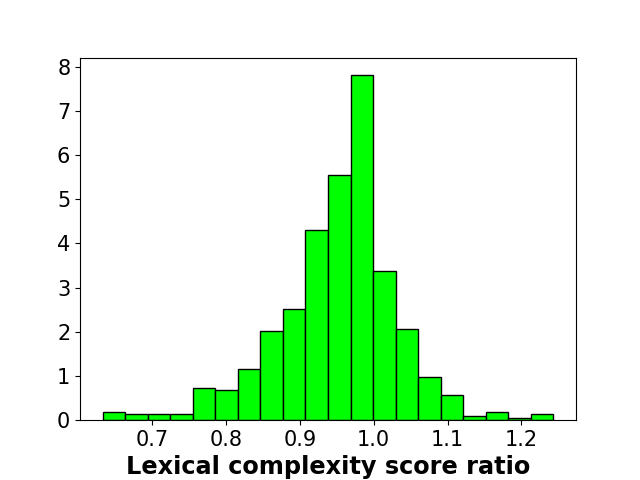}
        \end{minipage}
        \begin{minipage}{0.24\linewidth}
            \centering
            \includegraphics[width=1.1\linewidth]{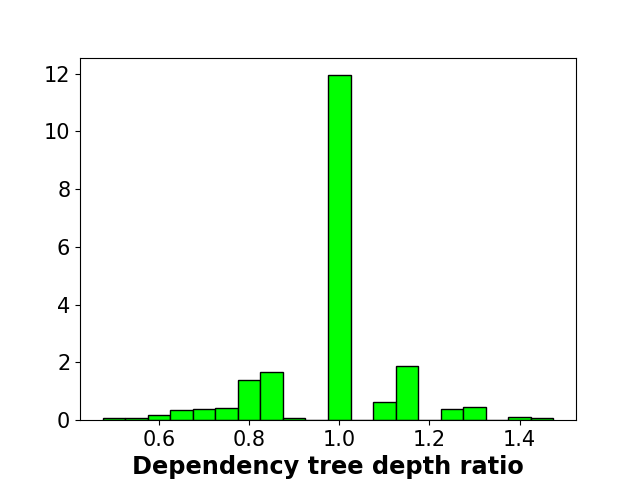}
        \end{minipage}
    \caption{Density of text features in simplifications from CSS.}
    \label{figure2}
\end{figure*}

Figure \ref{figure2} shows the density histograms of the features of CSS except \textbf{Number of sentence splits} and \textbf{Word deletion only}. 
For some key features that significantly demonstrate the difference between Chinese and English SS datasets, we highlight these statistics as percentages in Table \ref{table4}, and report the statistics of ASSET as a comparison.

Nearly half of the instances in CSS perform more than one simplification operation, which shows the diversity of simplification operations in CSS. By analyzing the replace-only Levenshtein distance and the proportion of words deleted/added, we can see how much the annotators have paraphrased the original sentence. Both CSS and ASSET have a very low ratio of word deletion only, which means that few extractive compression operations were performed.

Sentence splitting is a common operation in English sentence simplification. Although annotators in ASSET tended to not split sentences, they still performed sentence splitting at a rate of 20.2\% \cite{asset}, and this rate can even reach 68.2\% in HSplit. However, the percentage of sentence splitting is only 11.8 in CSS. A reasonable explanation for this phenomenon is that complex English sentences usually contain many nested clauses. Annotators may simplify these sentences by splitting the clause. And a complex Chinese sentence is usually constituted by many short sentences instead of nested clauses. This explanation is complemented by the distributions of dependency tree depth and the percentage of reordered words. In summary,\textbf{ Chinese SS do fewer structural changes than English.}

We introduce compression operation to simplify sentences in CSS, same with ASSET. However, Table \ref{table4} shows that the compression ratio on CSS is much lower than on ASSET. CSS has a high density of a compression ratio of 1.0, even has many instances with compression levels greater than 1.0. This phenomenon can be explained by the frequent use of idioms. In Chinese, an idiom often alludes to a story or historical quotation,
compressing a lot of information. It is difficult to simplify idioms just by replacing words. Annotators usually used a short sentence to explain an idiom, which leads to the above phenomenon.  \textbf{The lexical simplification of Chinese is different from English because of the existence of idioms.}

\subsection{Human Rating of CSS} \label{human rating}
In this section, we measure the quality of the CSS dataset using human judges. Workers needed to satisfy the same basic requirements as described in Section \ref{Data Collection}, and passed the Qualification Test that was designed for human evaluation. Following \citet{asset}, we rated the quality of simplifications based on three criteria: simplicity, fluency (or grammaticality), and meaning. Simplicity is the most important indicator in this task.

We invited three workers to evaluate the quality of the CSS dataset with the above criteria. We randomly chose 100 original sentences from the dataset and, for each of them, we sampled one manual simplification. Workers were asked to use the five-point Likert scale to submit their level of agreement (1:Strongly disagree, 5:Strongly agree) with the following statements:

\begin{itemize}
\item[$\bullet$] \textbf{Simplicity}: 
The simplified sentence is simpler and easier to understand than the original sentence.

\item[$\bullet$] \textbf{Fluency}:
The simplified sentence is fluent and free of grammatical errors.

\item[$\bullet$] \textbf{Meaning}:
The simplified sentence adequately preserves the meaning of the original, perhaps omitting the least important information.
\end{itemize}

The average simplicity score is 3.88, indicating the simplification of the original sentence is good enough. The average fluency scores reach 4.83, probably because the simplified sentences are written by humans and are easy to read. 
The meaning score achieves 4.71, implying that the references express the meaning of the original sentences precisely.
In all, the quality of CSS is guaranteed.

\subsection{Additional Dataset for Few-shot Setting} \label{additional}
We annotated an additional dataset following the annotation process described in Section \ref{Data Collection}. Different from CSS, this dataset only consists of 288 manual simplifications, with only one reference for each original sentence. We use part of the dataset as the validation set in our experiment.

The ability to efficiently learn from limited data is critical for NLP tasks. Recently, zero- and few-shot learning with large-scale pre-trained language models have achieved promising progress on generative tasks \cite{vu2022overcoming}. We released this additional dataset to facilitate future works on few-shot sentence simplification. Researchers are free to split the dataset into training and validation sets, using them in the few-shot scenario, and evaluate on CSS.

\section{Experiments}
In this section, we conducted a series of experiments to explore how to train Chinese SS models in low-resource scenarios. Some of them could serve as the baselines for future studies. All the models are tested on CSS. 


Following \citet{muss}, We first implement the following simple methods for comparison.

\paragraph{Identity}
It simply outputs the original sentence, which means the original sentence and the simplification are exactly the same.

\paragraph{Truncation}
The original sentence is truncated and only the first 80 percent of words are retained.
 
\paragraph{Gold Reference}
We report gold reference scores in CSS as two references are available. We compute scores of one manual simplification, using another simplification as a reference. The scores are then averaged over all references.

We then introduce several unsupervised methods and zero/few-short methods for comparison. 
\subsection{Unsupervised Method} \label{unsupervised}
Automatic SS systems rely on unsupervised techniques when supervised training data is unavailable.
\citet{NMT} proposed an unsupervised simplification\footnote{Previous works \cite{muss} used the term \emph{unsupervised simplification} to describe works that do not use any labeled parallel simplification data While using some supervised components.} method to build a SS parallel corpus based on a large-scale bilingual translation corpus, which has achieved state-of-the-art results in multiple languages. We replicated this current unsupervised state-of-the-art model in Chinese.

Specifically, we chose English as the bridge language and used News-commentary \textit{en-zh} dataset\footnote{News-commentary is a common dataset in the field of neural machine translation, we download it from \url{https://data.statmt.org/news-commentary/v15/training/}.} as the high-resource bilingual translation corpus.
Then, we used a machine translation model\footnote{\url{https://huggingface.co/Helsinki-NLP/opus-mt-en-zh}} to translate the English sentences to Chinese. The source sentences (Chinese) and the translated sentences constituted pseudo-complex-simple sentence pairs. Different from the original work \cite{NMT}, We filtered pseudo-SS corpus only by BLEU because FKGL metric is not suitable for Chinese.

To compare with the model trained by pseudo-SS data, We provide \textbf{translate training} that the original sentence and simplification all are translated from the English WikiLarge dataset as a baseline. We use the same translation model and data size to make a fair comparison.

\subsection{Zero- and Few-shot Transfer}
In addition to unsupervised methods, recent works on zero- and few-shot learning with pre-trained language models can provide a potential solution for performing the SS task in a low-resource language \cite{mallinson2020zero}. We conduct experiments to explore whether the model can obtain prior SS knowledge through \textbf{cross-lingual} transfer and \textbf{cross-task} transfer. 
And all the models are trained on mT5 \cite{mt5}, a variant of T5 that was pre-trained on a multilingual dataset.
\paragraph{Wikilarge Zero-shot Transfer}
We finetuned mT5 using Wikilarge \cite{wikilarge} dataset, and then applied the model to conduct the Chinese SS task. This experiment attempts to transfer knowledge from rich-resource language to low-resource language, leveraging the powerful cross-lingual transfer ability of mT5.

\paragraph{LCSTS Zero-shot Transfer}
LCSTS \cite{hu2015lcsts} is a Chinese short text summarization dataset. The tasks of sentence simplification and summarization both need the ability of compression. We trained mT5 with LCSTS and tested it on CSS, attempting to transfer knowledge from a similar task.

We also report the results of \textbf{few-shot transfer}, which continues to finetune above models with the additional dataset we have described in Section \ref{additional}. A \textbf{few-shot baseline} that is directly finetuned with the same additional dataset but without any additional training is provided to compare with these few-shot models.

Please refer to Appendix \ref{training setting} for the training details of the above models.

\begin{table*}[t]
\centering
\small
\setlength{\tabcolsep}{7mm}{
\begin{tabular}{@{}lccc@{}}
\hline
\multicolumn{1}{l|}{}  &\multicolumn{3}{c}{\textbf{CSS}} \\
\multicolumn{1}{l|}{} &\multicolumn{1}{c}{SARI$_{char}$} & SARI$_{word}$ & \multicolumn{1}{l}{BLEU} \\ \hline
\multicolumn{4}{l}{\textit{Baselines and Gold Reference}} \\  \hline
\multicolumn{1}{l|}{Identity Baseline} & \multicolumn{1}{c}{29.08}  & \multicolumn{1}{c}{27.61}  & \multicolumn{1}{c}{88.77}    \\
\multicolumn{1}{l|}{Truncation Baseline} &\multicolumn{1}{c}{32.95}  &\multicolumn{1}{c}{33.18}  &\multicolumn{1}{c}{76.36}   \\
\multicolumn{1}{l|}{Gold Reference} &\multicolumn{1}{c}{46.72}  &\multicolumn{1}{c}{45.71}  &\multicolumn{1}{c}{65.31}  \\ \hline
\multicolumn{4}{l}{\textit{Unsupervised Method}} \\  \hline
\multicolumn{1}{l|}{\citet{NMT}} &\multicolumn{1}{c}{\underline{36.27}}  &\multicolumn{1}{c}{33.39}  &\multicolumn{1}{c}{63.47}   \\
\multicolumn{1}{l|}{Translate Training} &\multicolumn{1}{c}{36.02}  &\multicolumn{1}{c}{\underline{34.44}}  &\multicolumn{1}{c}{71.41}  \\ \hline
\multicolumn{4}{l}{\textit{Zero- and Few-shot Transfer}} \\ \hline
\multicolumn{1}{l|}{Wikilarge Zero-shot Transfer } &\multicolumn{1}{c}{35.38}  &\multicolumn{1}{c}{33.92}  &\multicolumn{1}{c}{72.00}  \\
\multicolumn{1}{l|}{LCSTS  Zero-shot Transfer} &\multicolumn{1}{c}{22.34}  &\multicolumn{1}{c}{20.04}  &\multicolumn{1}{c}{20.77}    \\
\multicolumn{1}{l|}{Few-shot Baseline} &\multicolumn{1}{c}{\textbf{37.57}}  &\multicolumn{1}{c}{\textbf{35.97}}  &\multicolumn{1}{c}{\textbf{74.71}}  \\
\multicolumn{1}{l|}{Wikilarge Few-shot Transfer} &\multicolumn{1}{c}{35.59}  &\multicolumn{1}{c}{34.10}  &\multicolumn{1}{c}{\underline{72.13}}  \\
\multicolumn{1}{l|}{LCSTS Few-shot Transfer} &\multicolumn{1}{c}{34.23}  &\multicolumn{1}{c}{32.27}  &\multicolumn{1}{c}{64.08}   \\  \hline
\end{tabular}}
\caption{The automatic evaluation results on CSS test set. We use \textbf{Bold} to mark the best result and \underline{underline} the second-best result. SARI$_{char}$ means the value of SARI at character level, and SARI$_{word}$ means the value of SARI at word level.}
\label{automatic result}
\end{table*}

\section{Evaluation Results}\label{experiment result}

\subsection{Automatic Evaluation Results}
We use SARI and BLEU, standard metrics that were widely used in previous English sentence simplification work, to evaluate our Chinese models.

\paragraph{SARI}
The most commonly used automatic evaluation metric for sentence simplification is the SARI \cite{xu2016sari} metric, which compares the output of simplification systems with the original sentence and gold references to measure the simplicity gain.
The correlation of SARI with human judgments of simplicity proved to be high \cite{hsplit}. We compute SARI with the \textit{EASSE} simplification evaluation suite \cite{easse}.\footnote{We use the latest version of SARI implemented in \textit{EASSE} which fixes bugs and inconsistencies from the traditional implementation.}
In our experiments, We report SARI at both the character level and word level, which means two different tokenize ways of processing Chinese text.
\paragraph{BLEU}
BLEU \cite{bleu} is a metric to measure the similarity between the system outputs and the human references, which relies on the number of n-grams in the output that match n-grams in the references, independently of position. We calculate BLEU using \textit{NLTK} \cite{nltk}.

Previous work used FKGL \cite{fkgl} to measure the readability of the text. FKGL was tailored to be used in English only, and we do not report it in our experiments. A metric that can measure the readability of Chinese text is urgently needed in the research of Chinese sentence simplification. 

Table \ref{automatic result} shows the automatic evaluation results. The few-shot baseline exhibits surprising results, even surpassing other few-shot models. According to the results, the data of the English SS task and short text summarization task failed to provide additional improvement for few-shot Chinese SS. We, same with \citet{vu2022overcoming}, have observed severe catastrophic forgetting when we perform cross-lingual transfer with model tuning. Perhaps model tuning is not a good option for zero- and few-shot cross-lingual transfer. Through the previous data analysis in Section \ref{text features}, we can see that the ability of compression is not particularly important for Chinese SS, and adapting to the summarization task in advance even can harm the performance.
According to the ablation experiments in \citet{NMT}, building the pseudo corpus without FKGL selector can severely harm the performance of the model. This conclusion can explain why the unsupervised methods perform worse than expected.

\subsection{Human Evaluation Results}
We only chose three models to manually rate their outputs due to the high cost of human evaluation, which were selected based on the SARI metric on CSS.
We performed human evaluation according to the method described in Section \ref{human rating}. To maintain consistency, we chose the same 100 original sentences from CSS that were randomly selected for evaluating the quality of the dataset in Section \ref{human rating}. The human evaluation results are shown in Table \ref{table6}.

The few-shot baseline obtains the best result on the SARI score and the BLEU score. However, its simplicity score is only 1.66. The few-shot baseline can not be trained adequately with a small-scale dataset. In that scenario, the model tends to replicate original sentences only with very few modifications. Therefore, this model obtains a high score both on fluency and meaning. The Translate Training model gets the lowest simplicity score, demonstrating that machine translation systems fail to bridge the gap between Chinese and English SS tasks.
The human evaluation results show that we can not assess the performance of Chinese SS systems only by the value of SARI metric.

\begin{table}[]
\centering
\small
\setlength{\tabcolsep}{2mm}{
\begin{tabular}{lccc}
\hline
 & \multicolumn{1}{l}{Fluency} & \multicolumn{1}{l}{Meaning} & \multicolumn{1}{l}{Simplicity} \\ \hline
Gold Reference & 4.83 & 4.71 & 3.88 \\
\citet{NMT} & 4.43 & 4.06 & 2.76 \\
Translate Training & 4.67 & 4.49 & 1.59 \\
Few-shot Baseline & 4.62 & 4.67 & 1.66 \\ \hline
 & \multicolumn{1}{l}{} & \multicolumn{1}{l}{} & \multicolumn{1}{l}{}
\end{tabular}}
\caption{The result of human evaluation.}
\label{table6}
\end{table}

\subsection{Correlation of Automatic Metrics with Human Ratings}
We computed instance-level Spearman’s rank correlation coefficient \cite{spearman} between the automatic metrics (BLEU and SARI) and human ratings, using CSS as references. Results are reported in Table \ref{table7}. 

The SARI$_{char}$ metric has the highest correlation with the simplicity indicator, surpassing both BLEU and SARI$_{word}$. SARI$_{word}$ also shows a moderate positive correlation with simplicity. In terms of fluency and meaning, correlations are positive but low for both SARI metrics. BLEU is dubious to apply in the evaluation of SS \cite{xu2016sari,sulem2018semantic}, but also shows a low positive correlation with simplicity judgments when using CSS as references. In brief, the SARI metric is not very reliable but can still be used for reference when evaluating Chinese SS, and we need more metrics to reflect the quality of the outputs in different aspects.

\begin{table}[h]
\centering
\small
\setlength{\tabcolsep}{2mm}{
\begin{tabular}{lccc}
\hline
Metric & \multicolumn{1}{l}{Fluency} & \multicolumn{1}{l}{Meaning} & \multicolumn{1}{l}{Simplicity} \\ \hline
SARI$_{char}$ & 0.17 & 0.25 & 0.46 \\
SARI$_{word}$ & 0.18 & 0.26 & 0.41 \\
BLEU & 0.31 & 0.33 & 0.30 \\ \hline
\end{tabular}}
\caption{Spearman’s rank correlation of human ratings with automatic metrics.}
\label{table7}
\end{table}

\begin{table*}[t]
\centering
\small
\setlength{\tabcolsep}{7mm}{
\begin{tabular}{@{}lcccc@{}}
\hline
\multicolumn{2}{l|}{} & \multicolumn{3}{c}{\textbf{CSS}} \\
\multicolumn{2}{l|}{} & \multicolumn{1}{c}{SARI$_{char}$} & SARI$_{word}$ & \multicolumn{1}{l}{BLEU} \\  \hline
\multicolumn{5}{l}{\textit{GPT-3.5-turbo-0301}} \\ \hline
\multicolumn{2}{l|}{Zero-shot} &\multicolumn{1}{c}{31.95}  &\multicolumn{1}{c}{28.92}  &\multicolumn{1}{c}{42.22}    \\
\multicolumn{2}{l|}{Few-shot} &\multicolumn{1}{c}{\textbf{39.32}}  &\multicolumn{1}{c}{\textbf{36.57}}  &\multicolumn{1}{c}{60.67}   \\
\hline
\multicolumn{5}{l}{\textit{Vicuna-13B}} \\  \hline
\multicolumn{2}{l|}{Zero-shot} &\multicolumn{1}{c}{23.14}  &\multicolumn{1}{c}{20.67}  &\multicolumn{1}{c}{23.16} \\
\multicolumn{2}{l|}{Few-shot} &\multicolumn{1}{c}{28.68}  &\multicolumn{1}{c}{26.56}  &\multicolumn{1}{c}{38.04}  \\ \hline
\multicolumn{5}{l}{\textit{ChatGLM-6B}} \\  \hline
\multicolumn{2}{l|}{Zero-shot} &\multicolumn{1}{c}{35.17}  &\multicolumn{1}{c}{32.69}  &\multicolumn{1}{c}{56.59} \\
\multicolumn{2}{l|}{Few-shot} &\multicolumn{1}{c}{37.74}  &\multicolumn{1}{c}{35.70}  &\multicolumn{1}{c}{\textbf{66.37}}   \\
\hline
\end{tabular}}
\caption{The automatic evaluation results of LLMs on CSS test set. We use \textbf{Bold} to mark the best result.}
\label{LLMs_result}
\end{table*}

\section{Chinese Sentence Simplification via Large Language Models}
Large Language Models (LLMs) have demonstrated incredible ability in many natural language generation tasks and real-world applications. Recent research on sentence simplification has shown LLMs outperform state-of-the-art sentence simplification methods and generalize well across various low-resource languages \cite{feng2023sentence}. In this section, we select some representative large models and conduct zero-/few-shot experiments to evaluate them on CSS. We hope these results can supplement previous research on the cross-lingual SS capability of LLMs and serve as baselines for future studies.

According to the experiment result of \citet{sun2023teaching}, LLaMA seems unable to understand the prompt for simplifying sentences. Therefore, we choose those LLMs that can follow Chinese instructions after instruction tuning.

\paragraph{GPT-3.5-turbo-0301\footnote{\url{https://platform.openai.com/docs/models/gpt-3-5}}} Snapshot of GPT-3.5-turbo from March 1st 2023. Unlike GPT-3.5-turbo, this model will not receive updates. We choose this stable version to ensure reproducibility. 
\paragraph{Vicuna-13B\footnote{\url{https://github.com/lm-sys/FastChat}}} an open-source LLM trained by fine-tuning LLaMA \cite{2023llama} on 70K user-shared conversations collected from ShareGPT.
\paragraph{ChatGLM-6B\footnote{\url{https://github.com/THUDM/ChatGLM-6B}}} an open-source LLM based on General Language Model (GLM) framework \cite{du2022glm}, follows the training process similar to ChatGPT, optimized for Chinese QA and dialogue.

Then, we prompt LLMs to perform the Chinese SS task with zero-/few-shot style.
In the few-shot setting, we randomly choose 5 original-simplified sentence pairs from the additional dataset as demonstrations.
Please refer to Appendix \ref{prompt template} for our zero-/few-shot SS prompt template.

\subsection{Analysis and Discussion}
Table \ref{LLMs_result} shows the automatic evaluation result of LLMs.

Few-shot Baselines achieve better performance than Zero-shot Baselines for each LLM, which conform with our expectations. And ChatGLM performs better than Vicuna with only half the parameters. This is probably because Vicuna has not been trained specifically on the Chinese instruction dataset, although it showed Chinese capability to some extent. In fact, we even find some English characters in the simplification results of Vicuna.

It is slightly strange that GPT-3.5-turbo performs worse than ChatGLM in the zero-shot setting. There may exist a misalignment between GPT-3.5-turbo and human annotators on how to simplify Chinese sentences, and ChatGLM aligns well with humans through additional optimization for Chinese. However, GPT-3.5-turbo surpasses ChatGLM in the few-shot setting with powerful in-context learning ability, and outperforms all of baselines we report in Table \ref{automatic result}.

The above results illustrate that LLMs like GPT-3.5-turbo can serve as a high-quality Chinese sentence simplification system.

\section{Conclusion}
In this paper, we are committed to facilitating research on Chinese sentence simplification.
We introduced CSS, a new dataset for the evaluation of Chinese SS models. Simplifications in CSS were manually written by human annotators, and the simplification operations are also labeled. We give an in-depth analysis of CSS and reveal its characteristics. We further develop several unsupervised and zero/few-short methods on the dataset, and the experiment results can benefit future research.

\section*{Limitations}

We only create a high-quality dataset for evaluation and a small-scale dataset for few-shot learning, since the lack of a large-scale Chinese parallel SS corpus. The available research methods for Chinese SS are limited to unsupervised learning and few-shot learning. We hope a large-scale Chinese parallel SS corpus can be released in the future. Then we can directly train more supervised models for Chinese SS.

Furthermore, we only analyze whether the current standard metrics are suitable for the evaluation of Chinese SS, and leave the work of proposing a new metric for future study.
Due to time constraints, we do not perform a human evaluation for the output of LLMs. We hope to conduct a more comprehensive evaluation for LLMs in the future.

\section*{Ethics Statement}
We choose original sentences from the PFR corpus, which has been released to the public. There are no intellectual property disputes for our data source. All simplifications are collected from workers we employ, and adhere to the relevant code of ethics. We pay annotators a fair salary that matches their workload.

\section*{Acknowledgements}
This work was supported by National Key R\&D Program of China (2021YFF0901502), National Science Foundation of China (No.62161160339), State Key Laboratory of Media Convergence Production Technology and Systems and Key Laboratory of Science, Technology and Standard in Press Industry (Key Laboratory of Intelligent Press Media Technology). We appreciate the anonymous reviewers for their helpful comments. Xiaojun Wan is the corresponding author.

\bibliography{anthology,custom}

\begin{thebibliography}{55}
\expandafter\ifx\csname natexlab\endcsname\relax\def\natexlab#1{#1}\fi

\bibitem[{Alu{\'\i}sio et~al.(2008)Alu{\'\i}sio, Specia, Pardo, Maziero, and
  Fortes}]{aluisio2008towards}
Sandra~M Alu{\'\i}sio, Lucia Specia, Thiago~AS Pardo, Erick~G Maziero, and
  Renata~PM Fortes. 2008.
\newblock Towards brazilian portuguese automatic text simplification systems.
\newblock In \emph{Proceedings of the eighth ACM symposium on Document
  engineering}, pages 240--248.

\bibitem[{Alva-Manchego et~al.(2020{\natexlab{a}})Alva-Manchego, Martin,
  Bordes, Scarton, Sagot, and Specia}]{asset}
Fernando Alva-Manchego, Louis Martin, Antoine Bordes, Carolina Scarton,
  Beno{\^\i}t Sagot, and Lucia Specia. 2020{\natexlab{a}}.
\newblock Asset: A dataset for tuning and evaluation of sentence simplification
  models with multiple rewriting transformations.
\newblock In \emph{Proceedings of the 58th Annual Meeting of the Association
  for Computational Linguistics}, pages 4668--4679.

\bibitem[{Alva-Manchego et~al.(2019)Alva-Manchego, Martin, Scarton, and
  Specia}]{easse}
Fernando Alva-Manchego, Louis Martin, Carolina Scarton, and Lucia Specia. 2019.
\newblock Easse: Easier automatic sentence simplification evaluation.
\newblock In \emph{Proceedings of the 2019 Conference on Empirical Methods in
  Natural Language Processing and the 9th International Joint Conference on
  Natural Language Processing (EMNLP-IJCNLP): System Demonstrations}, pages
  49--54.

\bibitem[{Alva-Manchego et~al.(2020{\natexlab{b}})Alva-Manchego, Scarton, and
  Specia}]{alva2020data}
Fernando Alva-Manchego, Carolina Scarton, and Lucia Specia. 2020{\natexlab{b}}.
\newblock Data-driven sentence simplification: Survey and benchmark.
\newblock \emph{Computational Linguistics}, 46(1):135--187.

\bibitem[{Aprosio et~al.(2019)Aprosio, Tonelli, Turchi, Negri, and
  Di~Gangi}]{aprosio}
Alessio~Palmero Aprosio, Sara Tonelli, Marco Turchi, Matteo Negri, and Mattia~A
  Di~Gangi. 2019.
\newblock Neural text simplification in low-resource conditions using weak
  supervision.
\newblock In \emph{Proceedings of the Workshop on Methods for Optimizing and
  Evaluating Neural Language Generation}, pages 37--44.

\bibitem[{Battisti et~al.(2020)Battisti, Pf{\"u}tze, S{\"a}uberli, Kostrzewa,
  and Ebling}]{battisti2020corpus}
Alessia Battisti, Dominik Pf{\"u}tze, Andreas S{\"a}uberli, Marek Kostrzewa,
  and Sarah Ebling. 2020.
\newblock A corpus for automatic readability assessment and text simplification
  of german.
\newblock In \emph{Proceedings of the 12th Language Resources and Evaluation
  Conference}, pages 3302--3311.

\bibitem[{Bird(2006)}]{nltk}
Steven Bird. 2006.
\newblock Nltk: the natural language toolkit.
\newblock In \emph{Proceedings of the COLING/ACL 2006 Interactive Presentation
  Sessions}, pages 69--72.

\bibitem[{Brunato et~al.(2015)Brunato, Dell’Orletta, Venturi, and
  Montemagni}]{brunato2015design}
Dominique Brunato, Felice Dell’Orletta, Giulia Venturi, and Simonetta
  Montemagni. 2015.
\newblock Design and annotation of the first italian corpus for text
  simplification.
\newblock In \emph{Proceedings of The 9th Linguistic Annotation Workshop},
  pages 31--41.

\bibitem[{Carroll et~al.(1998)Carroll, Minnen, Canning, Devlin, and
  Tait}]{carroll1998practical}
John Carroll, Guido Minnen, Yvonne Canning, Siobhan Devlin, and John Tait.
  1998.
\newblock Practical simplification of english newspaper text to assist aphasic
  readers.
\newblock In \emph{Proceedings of the AAAI-98 Workshop on Integrating
  Artificial Intelligence and Assistive Technology}, pages 7--10. Citeseer.

\bibitem[{De~Belder and Moens(2010)}]{de2010text}
Jan De~Belder and Marie-Francine Moens. 2010.
\newblock Text simplification for children.
\newblock In \emph{Prroceedings of the SIGIR workshop on accessible search
  systems}, pages 19--26. ACM; New York.

\bibitem[{Dmitrieva and Tiedemann(2021)}]{dmitrieva2021multi}
Anna Dmitrieva and J{\"o}rg Tiedemann. 2021.
\newblock A multi-task learning approach to text simplification.
\newblock In \emph{International Conference on Analysis of Images, Social
  Networks and Texts}, pages 78--89. Springer.

\bibitem[{Dmitrieva et~al.(2021)Dmitrieva, Tiedemann
  et~al.}]{RussianWithLearnerdata}
Anna Dmitrieva, J{\"o}rg Tiedemann, et~al. 2021.
\newblock Creating an aligned russian text simplification dataset from language
  learner data.
\newblock In \emph{Proceedings of the 8th Workshop on Balto-Slavic Natural
  Language Processing}. ACL Anthology.

\bibitem[{Dong et~al.(2019)Dong, Li, Rezagholizadeh, and
  Cheung}]{dong2019editnts}
Yue Dong, Zichao Li, Mehdi Rezagholizadeh, and Jackie Chi~Kit Cheung. 2019.
\newblock Editnts: An neural programmer-interpreter model for sentence
  simplification through explicit editing.
\newblock In \emph{Proceedings of the 57th Annual Meeting of the Association
  for Computational Linguistics}, pages 3393--3402.

\bibitem[{Du et~al.(2022)Du, Qian, Liu, Ding, Qiu, Yang, and Tang}]{du2022glm}
Zhengxiao Du, Yujie Qian, Xiao Liu, Ming Ding, Jiezhong Qiu, Zhilin Yang, and
  Jie Tang. 2022.
\newblock Glm: General language model pretraining with autoregressive blank
  infilling.
\newblock In \emph{Proceedings of the 60th Annual Meeting of the Association
  for Computational Linguistics (Volume 1: Long Papers)}, pages 320--335.

\bibitem[{Evans et~al.(2014)Evans, Or{\u{a}}san, and
  Dornescu}]{evans2014evaluation}
Richard Evans, Constantin Or{\u{a}}san, and Iustin Dornescu. 2014.
\newblock \href {https://doi.org/10.3115/v1/W14-1215} {An evaluation of
  syntactic simplification rules for people with autism}.
\newblock In \emph{Proceedings of the 3rd Workshop on Predicting and Improving
  Text Readability for Target Reader Populations ({PITR})}, pages 131--140,
  Gothenburg, Sweden. Association for Computational Linguistics.

\bibitem[{Feng(2008)}]{feng2008text}
Lijun Feng. 2008.
\newblock Text simplification: A survey.
\newblock \emph{The City University of New York, Technical Report}.

\bibitem[{Feng et~al.(2023)Feng, Qiang, Li, Yuan, and Zhu}]{feng2023sentence}
Yutao Feng, Jipeng Qiang, Yun Li, Yunhao Yuan, and Yi~Zhu. 2023.
\newblock Sentence simplification via large language models.
\newblock \emph{arXiv preprint arXiv:2302.11957}.

\bibitem[{Gala et~al.(2020)Gala, Tack, Javourey-Drevet, Fran{\c{c}}ois, and
  Ziegler}]{gala2020alector}
N{\'u}ria Gala, Ana{\"\i}s Tack, Ludivine Javourey-Drevet, Thomas
  Fran{\c{c}}ois, and Johannes~C Ziegler. 2020.
\newblock Alector: A parallel corpus of simplified french texts with alignments
  of misreadings by poor and dyslexic readers.
\newblock In \emph{Language Resources and Evaluation for Language Technologies
  (LREC)}.

\bibitem[{Hu et~al.(2015)Hu, Chen, and Zhu}]{hu2015lcsts}
Baotian Hu, Qingcai Chen, and Fangze Zhu. 2015.
\newblock Lcsts: A large scale chinese short text summarization dataset.
\newblock In \emph{Proceedings of the 2015 Conference on Empirical Methods in
  Natural Language Processing}, pages 1967--1972.

\bibitem[{Kajiwara and Komachi(2018)}]{kajiwara2018text}
Tomoyuki Kajiwara and M~Komachi. 2018.
\newblock Text simplification without simplified corpora.
\newblock \emph{The Journal of Natural Language Processing}, 25:223--249.

\bibitem[{Kajiwara et~al.(2013)Kajiwara, Matsumoto, and
  Yamamoto}]{kajiwara2013selecting}
Tomoyuki Kajiwara, Hiroshi Matsumoto, and Kazuhide Yamamoto. 2013.
\newblock Selecting proper lexical paraphrase for children.
\newblock In \emph{Proceedings of the 25th Conference on Computational
  Linguistics and Speech Processing (ROCLING 2013)}, pages 59--73.

\bibitem[{Katsuta and Yamamoto(2019)}]{katsuta2019}
Akihiro Katsuta and Kazuhide Yamamoto. 2019.
\newblock Improving text simplification by corpus expansion with unsupervised
  learning.
\newblock In \emph{2019 International Conference on Asian Language Processing
  (IALP)}, pages 216--221. IEEE.

\bibitem[{Kincaid et~al.(1975)Kincaid, Fishburne~Jr, Rogers, and
  Chissom}]{fkgl}
J~Peter Kincaid, Robert~P Fishburne~Jr, Richard~L Rogers, and Brad~S Chissom.
  1975.
\newblock Derivation of new readability formulas (automated readability index,
  fog count and flesch reading ease formula) for navy enlisted personnel.
\newblock Technical report, Naval Technical Training Command Millington TN
  Research Branch.

\bibitem[{Kingma and Ba(2014)}]{adam}
Diederik~P Kingma and Jimmy Ba. 2014.
\newblock Adam: A method for stochastic optimization.
\newblock \emph{arXiv preprint arXiv:1412.6980}.

\bibitem[{Kumar et~al.(2020)Kumar, Mou, Golab, and Vechtomova}]{kumariterative}
Dhruv Kumar, Lili Mou, Lukasz Golab, and Olga Vechtomova. 2020.
\newblock Iterative edit-based unsupervised sentence simplification.
\newblock In \emph{Proceedings of the 58th Annual Meeting of the Association
  for Computational Linguistics}, pages 7918--7928.

\bibitem[{Levenshtein et~al.(1966)}]{levenshtein1966binary}
Vladimir~I Levenshtein et~al. 1966.
\newblock Binary codes capable of correcting deletions, insertions, and
  reversals.
\newblock In \emph{Soviet physics doklady}, volume~10, pages 707--710. Soviet
  Union.

\bibitem[{Li et~al.(2022)Li, Li, Zhang, Li, Wei, Cui, and Wang}]{li2022c3kg}
Dawei Li, Yanran Li, Jiayi Zhang, Ke~Li, Chen Wei, Jianwei Cui, and Bin Wang.
  2022.
\newblock C3kg: A chinese commonsense conversation knowledge graph.
\newblock In \emph{Findings of the Association for Computational Linguistics:
  ACL 2022}, pages 1369--1383.

\bibitem[{Lu et~al.(2021)Lu, Qiang, Li, Yuan, and Zhu}]{NMT}
Xinyu Lu, Jipeng Qiang, Yun Li, Yunhao Yuan, and Yi~Zhu. 2021.
\newblock An unsupervised method for building sentence simplification corpora
  in multiple languages.
\newblock In \emph{Findings of the Association for Computational Linguistics:
  EMNLP 2021}, pages 227--237.

\bibitem[{Mallinson et~al.(2020)Mallinson, Sennrich, and
  Lapata}]{mallinson2020zero}
Jonathan Mallinson, Rico Sennrich, and Mirella Lapata. 2020.
\newblock Zero-shot crosslingual sentence simplification.
\newblock In \emph{Proceedings of the 2020 Conference on Empirical Methods in
  Natural Language Processing (EMNLP)}, pages 5109--5126.

\bibitem[{Martin et~al.(2020{\natexlab{a}})Martin, De~La~Clergerie, Sagot, and
  Bordes}]{ACCESS}
Louis Martin, {\'E}ric~Villemonte De~La~Clergerie, Beno{\^\i}t Sagot, and
  Antoine Bordes. 2020{\natexlab{a}}.
\newblock Controllable sentence simplification.
\newblock In \emph{Proceedings of the 12th Language Resources and Evaluation
  Conference}, pages 4689--4698.

\bibitem[{Martin et~al.(2020{\natexlab{b}})Martin, Fan, de~la Clergerie,
  Bordes, and Sagot}]{muss}
Louis Martin, Angela Fan, {\'E}ric de~la Clergerie, Antoine Bordes, and
  Beno{\^\i}t Sagot. 2020{\natexlab{b}}.
\newblock Muss: multilingual unsupervised sentence simplification by mining
  paraphrases.
\newblock \emph{arXiv preprint arXiv:2005.00352}.

\bibitem[{Omelianchuk et~al.(2021)Omelianchuk, Raheja, and Skurzhanskyi}]{tst}
Kostiantyn Omelianchuk, Vipul Raheja, and Oleksandr Skurzhanskyi. 2021.
\newblock Text simplification by tagging.
\newblock In \emph{Proceedings of the 16th Workshop on Innovative Use of NLP
  for Building Educational Applications}, pages 11--25.

\bibitem[{Paetzold and Specia(2016{\natexlab{a}})}]{paetzold2016semeval}
Gustavo Paetzold and Lucia Specia. 2016{\natexlab{a}}.
\newblock Semeval 2016 task 11: Complex word identification.
\newblock In \emph{Proceedings of the 10th International Workshop on Semantic
  Evaluation (SemEval-2016)}, pages 560--569.

\bibitem[{Paetzold and Specia(2016{\natexlab{b}})}]{paetzold2016unsupervised}
Gustavo Paetzold and Lucia Specia. 2016{\natexlab{b}}.
\newblock Unsupervised lexical simplification for non-native speakers.
\newblock In \emph{Proceedings of the AAAI Conference on Artificial
  Intelligence}, volume~30.

\bibitem[{Papineni et~al.(2002)Papineni, Roukos, Ward, and Zhu}]{bleu}
Kishore Papineni, Salim Roukos, Todd Ward, and Wei-Jing Zhu. 2002.
\newblock Bleu: a method for automatic evaluation of machine translation.
\newblock In \emph{Proceedings of the 40th annual meeting of the Association
  for Computational Linguistics}, pages 311--318.

\bibitem[{Paszke et~al.(2019)Paszke, Gross, Massa, Lerer, Bradbury, Chanan,
  Killeen, Lin, Gimelshein, Antiga et~al.}]{pytorch}
Adam Paszke, Sam Gross, Francisco Massa, Adam Lerer, James Bradbury, Gregory
  Chanan, Trevor Killeen, Zeming Lin, Natalia Gimelshein, Luca Antiga, et~al.
  2019.
\newblock Pytorch: An imperative style, high-performance deep learning library.
\newblock \emph{Advances in neural information processing systems}, 32.

\bibitem[{Petersen and Ostendorf(2007)}]{petersen2007text}
Sarah~E Petersen and Mari Ostendorf. 2007.
\newblock Text simplification for language learners: a corpus analysis.
\newblock In \emph{Workshop on speech and language technology in education}.
  Citeseer.

\bibitem[{Rello et~al.(2013)Rello, Bayarri, G{\'o}rriz, Baeza-Yates, Gupta,
  Kanvinde, Saggion, Bott, Carlini, and Topac}]{rello2013dyswebxia}
Luz Rello, Clara Bayarri, Azuki G{\'o}rriz, Ricardo Baeza-Yates, Saurabh Gupta,
  Gaurang Kanvinde, Horacio Saggion, Stefan Bott, Roberto Carlini, and Vasile
  Topac. 2013.
\newblock Dyswebxia 2.0! more accessible text for people with dyslexia.
\newblock In \emph{Proceedings of the 10th International Cross-Disciplinary
  Conference on Web Accessibility}, pages 1--2.

\bibitem[{Saggion et~al.(2015)Saggion, {\v{S}}tajner, Bott, Mille, Rello, and
  Drndarevic}]{saggion2015making}
Horacio Saggion, Sanja {\v{S}}tajner, Stefan Bott, Simon Mille, Luz Rello, and
  Biljana Drndarevic. 2015.
\newblock Making it simplext: Implementation and evaluation of a text
  simplification system for spanish.
\newblock \emph{ACM Transactions on Accessible Computing (TACCESS)},
  6(4):1--36.

\bibitem[{Sakhovskiy et~al.(2021)Sakhovskiy, Izhevskaya, Pestova, Tutubalina,
  Malykh, Smurov, and Artemova}]{RuEval}
Andrey Sakhovskiy, Alexandra Izhevskaya, Alena Pestova, Elena Tutubalina,
  Valentin Malykh, Ivan Smurov, and Ekaterina Artemova. 2021.
\newblock Rusimplesenteval-2021 shared task: evaluating sentence simplification
  for russian.
\newblock In \emph{Proceedings of the International Conference “Dialogue},
  pages 607--617.

\bibitem[{{\v{S}}tajner et~al.(2015){\v{S}}tajner, Calixto, and
  Saggion}]{vstajner2015automatic}
Sanja {\v{S}}tajner, Iacer Calixto, and Horacio Saggion. 2015.
\newblock Automatic text simplification for spanish: Comparative evaluation of
  various simplification strategies.
\newblock In \emph{Proceedings of the international conference recent advances
  in natural language processing}, pages 618--626.

\bibitem[{Sulem et~al.(2018{\natexlab{a}})Sulem, Abend, and Rappoport}]{hsplit}
Elior Sulem, Omri Abend, and Ari Rappoport. 2018{\natexlab{a}}.
\newblock Bleu is not suitable for the evaluation of text simplification.
\newblock In \emph{Proceedings of the 2018 Conference on Empirical Methods in
  Natural Language Processing}, pages 738--744.

\bibitem[{Sulem et~al.(2018{\natexlab{b}})Sulem, Abend, and
  Rappoport}]{sulem2018semantic}
Elior Sulem, Omri Abend, and Ari Rappoport. 2018{\natexlab{b}}.
\newblock Semantic structural evaluation for text simplification.
\newblock In \emph{Proceedings of the 2018 Conference of the North American
  Chapter of the Association for Computational Linguistics: Human Language
  Technologies, Volume 1 (Long Papers)}, pages 685--696.

\bibitem[{Sun et~al.(2023)Sun, Xu, and Wan}]{sun2023teaching}
Renliang Sun, Wei Xu, and Xiaojun Wan. 2023.
\newblock \href {http://arxiv.org/abs/2305.12463} {Teaching the pre-trained
  model to generate simple texts for text simplification}.

\bibitem[{Surya et~al.(2019)Surya, Mishra, Laha, Jain, and
  Sankaranarayanan}]{surya2019unsupervised}
Sai Surya, Abhijit Mishra, Anirban Laha, Parag Jain, and Karthik
  Sankaranarayanan. 2019.
\newblock Unsupervised neural text simplification.
\newblock In \emph{Proceedings of the 57th Annual Meeting of the Association
  for Computational Linguistics}, pages 2058--2068.

\bibitem[{Tonelli et~al.(2016)Tonelli, Aprosio, and
  Saltori}]{tonelli2016simpitiki}
Sara Tonelli, Alessio~Palmero Aprosio, and Francesca Saltori. 2016.
\newblock Simpitiki: a simplification corpus for italian.
\newblock In \emph{CLiC-it/EVALITA}.

\bibitem[{Touvron et~al.(2023)Touvron, Lavril, Izacard, Martinet, Lachaux,
  Lacroix, Rozi{\`e}re, Goyal, Hambro, Azhar et~al.}]{2023llama}
Hugo Touvron, Thibaut Lavril, Gautier Izacard, Xavier Martinet, Marie-Anne
  Lachaux, Timoth{\'e}e Lacroix, Baptiste Rozi{\`e}re, Naman Goyal, Eric
  Hambro, Faisal Azhar, et~al. 2023.
\newblock Llama: Open and efficient foundation language models.
\newblock \emph{arXiv preprint arXiv:2302.13971}.

\bibitem[{Vu et~al.(2022)Vu, Barua, Lester, Cer, Iyyer, and
  Constant}]{vu2022overcoming}
Tu~Vu, Aditya Barua, Brian Lester, Daniel Cer, Mohit Iyyer, and Noah Constant.
  2022.
\newblock Overcoming catastrophic forgetting in zero-shot cross-lingual
  generation.
\newblock \emph{arXiv preprint arXiv:2205.12647}.

\bibitem[{Wolf et~al.(2020)Wolf, Debut, Sanh, Chaumond, Delangue, Moi, Cistac,
  Rault, Louf, Funtowicz et~al.}]{wolf2020transformers}
Thomas Wolf, Lysandre Debut, Victor Sanh, Julien Chaumond, Clement Delangue,
  Anthony Moi, Pierric Cistac, Tim Rault, R{\'e}mi Louf, Morgan Funtowicz,
  et~al. 2020.
\newblock Transformers: State-of-the-art natural language processing.
\newblock In \emph{Proceedings of the 2020 conference on empirical methods in
  natural language processing: system demonstrations}, pages 38--45.

\bibitem[{Xu et~al.(2015)Xu, Callison-Burch, and Napoles}]{newsela}
Wei Xu, Chris Callison-Burch, and Courtney Napoles. 2015.
\newblock Problems in current text simplification research: New data can help.
\newblock \emph{Transactions of the Association for Computational Linguistics},
  3:283--297.

\bibitem[{Xu et~al.(2016)Xu, Napoles, Pavlick, Chen, and
  Callison-Burch}]{xu2016sari}
Wei Xu, Courtney Napoles, Ellie Pavlick, Quanze Chen, and Chris Callison-Burch.
  2016.
\newblock Optimizing statistical machine translation for text simplification.
\newblock \emph{Transactions of the Association for Computational Linguistics},
  4:401--415.

\bibitem[{Xue et~al.(2021)Xue, Constant, Roberts, Kale, Al-Rfou, Siddhant,
  Barua, and Raffel}]{mt5}
Linting Xue, Noah Constant, Adam Roberts, Mihir Kale, Rami Al-Rfou, Aditya
  Siddhant, Aditya Barua, and Colin Raffel. 2021.
\newblock mt5: A massively multilingual pre-trained text-to-text transformer.
\newblock In \emph{Proceedings of the 2021 Conference of the North American
  Chapter of the Association for Computational Linguistics: Human Language
  Technologies}, pages 483--498.

\bibitem[{Zhang and Lapata(2017)}]{wikilarge}
Xingxing Zhang and Mirella Lapata. 2017.
\newblock Sentence simplification with deep reinforcement learning.
\newblock \emph{arXiv preprint arXiv:1703.10931}.

\bibitem[{Zhu et~al.(2010)Zhu, Bernhard, and Gurevych}]{zhu2010monolingual}
Zhemin Zhu, Delphine Bernhard, and Iryna Gurevych. 2010.
\newblock A monolingual tree-based translation model for sentence
  simplification.
\newblock In \emph{Proceedings of the 23rd International Conference on
  Computational Linguistics (Coling 2010)}, pages 1353--1361.

\bibitem[{Zwillinger and Kokoska(1999)}]{spearman}
Daniel Zwillinger and Stephen Kokoska. 1999.
\newblock \emph{CRC standard probability and statistics tables and formulae}.
\newblock Crc Press.

\end{thebibliography}
\bibliographystyle{acl_natbib}

\appendix
\onecolumn
\section{Data source and Preprocessing}
\label{datasource}

\paragraph{Data Source}
 The PFR corpus (2014 version)\footnote{The PFR corpus are from \url{https://www.heywhale.com/mw/dataset}} was released by Peking University, including one year's (2014) newspaper material published by the People’s Daily. We chose the PFR corpus as the source of original sentences because People's Daily is the most authoritative and largest circulation newspaper in China, covering all aspects of social life.

\paragraph{Preprocessing}
The content of the corpus included the POS tag, and we restored articles to their original format and cut the text into sentences by punctuation. Then, we filtered out sentences with less than 30 tokens, since a short sentence may be difficult to simplify more.

\section{Simplification Examples}
\label{more_examples}
\begin{table*}[htbp]
\centering
\renewcommand{\arraystretch}{1.2}{
\begin{tabular}{ll}
\hline
Original &  \begin{CJK*}{UTF8}{gbsn}\footnotesize{\makecell[l]{随着深度学习、人工智能领域崛起，人类对于算力要求越来越高，显卡要想满足这些需求，\\必须不断迭代，但是随着摩尔定律失效，要想获得更高性能就需要更高的成本。\\With the rise of deep learning and artificial intelligence, human beings are demanding more and\\ more computing power, and graphics cards must continue to iterate for meeting these requirements, \\but with the failure of Moore's Law, to get higher performance will require higher costs.}}\end{CJK*}\\ \hline
Reference & \begin{CJK*}{UTF8}{gbsn}\footnotesize{\makecell[l]{随着深度学习、人工智能领域崛起，人类需要更高算力的显卡，但是这需要更高的成本。\\With the rise of deep learning and artificial intelligence, humans need graphics cards with\\ higher computing power, but this requires higher costs.}}\end{CJK*} \\ \hline
Operations &  Compression; Sentence paraphrasing\\ \hline
Original &  \begin{CJK*}{UTF8}{gbsn}\footnotesize{\makecell[l]{日侵略野心引起欧美警惕，在国际压力下日被迫放弃山东权益，从西伯利亚退兵。\\Japanese aggressive ambitions caused alarm in Europe and America, and under international \\ pressure Japan was forced to give up its rights in Shandong and retreat from Siberia.}}\end{CJK*}  \\ \hline
Reference &  \begin{CJK*}{UTF8}{gbsn}\footnotesize{\makecell[l]{日侵略野心引起欧美警惕。日本在国际压力下放弃山东权益，从西伯利亚退兵。\\Japanese aggressive ambitions caused alarm in Europe and America. Japan gave up its rights \\ in Shandong under international pressure and retreated from Siberia.}}\end{CJK*} \\ \hline
Operations & Compression; Sentence splitting\\
\hline
\end{tabular}}
\caption{Simplification examples with corresponding translations and operation tags in CSS.}
\end{table*}

\section{Detailed Training Settings}
\label{training setting}
We train all models in PyTorch \cite{pytorch}, and use the HuggingFace\footnote{\url{https://github.com/huggingface/}} \cite{wolf2020transformers} implementation of mT5 \cite{mt5}. We train each model on an A40 GPU. In all the experiments, we use the base version of mT5, and finetune it using Adam \cite{adam} optimizer with a batchsize of 32 and a warmup ratio of 0.1. Table \ref{table9} shows the detailed setting for our models. Besides, all few-shot transfer models are trained with the setting of \textbf{few-shot baseline} in the base of zero-shot transfer models.

\begin{table*}[h]
\centering
\begin{tabular}{@{}lccc@{}}
\hline
 & \multicolumn{1}{l}{Learning rate} & \multicolumn{1}{l}{Max-epoch} & \multicolumn{1}{l}{Eval steps} \\ \hline
\citet{NMT} & 2e-4 & 5 & 800 \\
Translate training & 2e-4 & 5 & 800 \\
Zero-shot Transfer model & 2e-4 & 5 & 800 \\
Few-shot Baseline & 1e-5 & 40 & 4 \\ \hline
\end{tabular}
\caption{Detailed hyperparameters of our model. Zero-shot Transfer model means \textbf{Wikilarge Zero-shot Transfer} and \textbf{LCSTS Zero-shot Transfer} in Table \ref{automatic result}.}
\label{table9}
\end{table*}

\newpage
\section{Prompt Template for Chinese Sentence Simplification}
\label{prompt template}
\begin{figure*}[h]
\centering
\fcolorbox{black}{gray!10}{\parbox{\linewidth}{
\center{\textcolor{red}{\textit{Zero-shot template}}}

\begin{CJK*}{UTF8}{gbsn}{\makecell[l]{请在保留原意的基础上简化以下句子：}}\end{CJK*} \\
\begin{CJK*}{UTF8}{gbsn}{\makecell[l]{原句：\color{blue}{\{Original Sentence\}}}}\end{CJK*} \\
\begin{CJK*}{UTF8}{gbsn}{\makecell[l]{简化句：\color{blue}{\{Outputs\}}}}\end{CJK*} \\
~\\

\center{\textcolor{red}{\textit{Few-shot template}}} \\
\begin{CJK*}{UTF8}{gbsn}{\makecell[l]{请在保留原意的基础上简化句子,以下是五个句子简化的示例：}}\end{CJK*} \\
\begin{CJK*}{UTF8}{gbsn}{\makecell[l]{原句：\color{blue}{\{Original Sentence\}}}}\end{CJK*} \\
\begin{CJK*}{UTF8}{gbsn}{\makecell[l]{简化句：\color{blue}{\{Simplified Sentence\}}}}\end{CJK*} \\
......\\
\begin{CJK*}{UTF8}{gbsn}{\makecell[l]{原句：\color{blue}{\{Original Sentence\}}}}\end{CJK*} \\
\begin{CJK*}{UTF8}{gbsn}{\makecell[l]{简化句：\color{blue}{\{Outputs\}}}}\end{CJK*} 
} 
}

\caption{The prompt template for zero-/few-shot Chinese sentence simplification}
\end{figure*}

\end{document}